\documentclass{article}
\PassOptionsToPackage{numbers, sort}{natbib}
\usepackage[final]{neurips_2020_ml4ps}
\usepackage[utf8]{inputenc} 
\usepackage[T1]{fontenc}    
\usepackage[hidelinks]{hyperref}       
\usepackage{url}            
\usepackage{booktabs}       
\usepackage{amsmath,amsfonts,amssymb}
\usepackage{nicefrac}       
\usepackage{microtype}
\usepackage[labelfont=bf]{caption}
\usepackage{xcolor}
\usepackage{mathtools}
\usepackage{float}
\usepackage{graphicx}
\usepackage{epstopdf}
\usepackage{etoolbox}
\apptocmd{\sloppy}{\hbadness 10000\relax}{}{}

\newcommand{\abs}[1]{\left|{#1}\right|}

\newcommand*{\pd}[3][]{{\frac{\partial^{#1} #2}{\partial #3}}}
\newcommand{\bracs}[1]{\left({#1}\right)}
\newcommand{\bracm}[1]{\left[{#1}\right]}
\newcommand{\bracl}[1]{\left\{{#1}\right\}}
\title{Multi-Constitutive Neural Network for Large Deformation Poromechanics Problem}
\author{%
  Qi Zhang$^{1\dagger}$, Yilin Chen$^{1\dagger}$, Ziyi Yang$^{2\dagger}$, Eric Darve$^{2, 3}$ \\
  $^1$ Department of Civil and Environmental Engineering, Stanford University \\
  $^2$ Department of Mechanical Engineering, Stanford University \\
  $^3$ Institute for Computational and Mathematical Engineering, Stanford University \\
  $^\dagger$ Equal Contribution \\
  \texttt{\{qzhang94, yilinc2, ziyi.yang, darve\}@stanford.edu}
}

\begin{document}

\maketitle

\begin{abstract}
In this paper, we study the problem of large-strain consolidation in poromechanics with deep neural networks (DNN). Given different material properties and different loading conditions, the goal is to predict pore pressure and settlement. We propose a novel method ``multi-constitutive neural network'' (MCNN) such that one model can solve several different constitutive laws. We introduce a one-hot encoding vector as an additional input vector, which is used to label the constitutive law we wish to solve. Then we build a DNN which takes $(\hat{X}, \hat{t})$ as input along with a constitutive law label and outputs the corresponding solution. It is the first time, to our knowledge, that we can evaluate multi-constitutive laws through only one training process while still obtaining good accuracies. We found that MCNN trained to solve multiple PDEs outperforms individual neural network solvers trained with PDE in some cases.
\end{abstract}



\section{Introduction}
\label{intro}

In the last decade, machine learning, especially deep neural networks (DNN) have yielded revolutionary results across diverse disciplines, and a fair amount of research has also been done related to differential equations that appear in many engineering fields. In the work done by Raissi et al. \cite{Raissi2017a,Raissi2017b,Raissi2019}, the physics-informed neural network (PINN) is proposed to solve a given physics problem described by a specific partial differential equation (PDE). However, there might exist multiple PDEs for a given problem, corresponding to choosing different constitutive laws. Traditionally, multiple models need to be built and trained to solve them separately, which might be computationally expensive and time-consuming. The multi-constitutive problem is especially important in large deformation poromechanics where many different constitutive laws which are derived from distinct forms of strain-energy density functions \cite{Holzapfel2000} can exist for one application \cite{Zheng2019}. In this paper, we propose a multi-constitutive neural network (MCNN); it uses an input vector which encodes the choice of constitutive law as an extra input to the network. Formally, let us index each constitutive law by some integer $i$. Denote $J_i(x,t)$ the solution to the PDE (or the sought after function) with constitutive law $i$, as a function of $x$ and $t$. Our neural network is a function $\tilde{J}(x,t,i; \theta)$, where $\theta$ are the usual DNN parameters (weights and biases), $i$ is the index of the constitutive law, and $\tilde{J}(x,t,i; \theta)$ is an approximation of the solution $J_i(x,t)$ for law $i$.

By solving the large-strain consolidation problem \cite{Xie2004,Gibson1967,Zheng2019,Selvadurai2016,Mac2016}, we have demonstrated the generalization capability of our model when incorporating multiple constitutive laws; our approach also requires fewer model parameters. The novel contributions of this paper are twofold: (1) A general deep learning framework MCNN that ensembles multi-constitutive laws and solves the resulting governing equations ``essentially simultaneously'' (simply by changing the input $i$, encoded as a one-hot vector); (2) Simultaneous training is more efficient and provides better generalization than individual training. In our benchmarks, for the second and the third cases, MCNN is more accurate than individual PINN models which are trained independently for each constitutive law.

\section{Multi-Constitutive Laws in Large Deformation Poromechanics}
\label{poro}

We briefly review the multi-constitutive laws used in the large-strain consolidation problem. In order to eliminate the effect of choice of units in data, we employ a non-dimensionalization technique, which means variables are described in their dimensionless form. In this paper, three constitutive laws of hyper-elasticity are investigated \cite{Holzapfel2000}, and here we give their forms after incorporating the momentum balance equation and effective stress principle \cite{Zheng2019,Coussy2004,Choo2018,Zhang2020b,Detournay1993}:

(i) Saint-Venant Kirchhoff law (law 1):
\begin{equation}
  \label{S-V_K}
  \pd{\hat{p}}{\hat{X}} = \frac{3J^2 - 1}{2} \pd{J}{\hat{X}}
\end{equation}
(ii) Modified Saint-Venant Kirchhoff law (law 2):
\begin{equation}
  \label{mS-V_K}
  \pd{\hat{p}}{\hat{X}} = \bracm{\hat{\gamma} \frac{1 - \log{J}}{J^2} + \hat{\mu} \bracs{3J^2 - 1}} \pd{J}{\hat{X}}
\end{equation}
(iii) Neo-Hookean law (law 3):
\begin{equation}
  \label{N-H}
  \pd{\hat{p}}{\hat{X}} = \bracm{\hat{\gamma} \frac{1 - \log{J}}{J^2} + \hat{\mu} \bracs{1 + \frac{1}{J^2}}} \pd{J}{\hat{X}}
\end{equation}
where in above three equations, $J$ is a measure of the deformation gradient, $\hat{p}$ is the dimensionless fluid (pore) pressure, $\hat{X} \in [0, 1]$ is the dimensionless coordinate, $\hat{\gamma}$ and $\hat{\mu}$ are the dimensionless Lam{\'{e}} constants. The mass balance equation is independent of multi-constitutive laws and is given by
\begin{equation}
  \label{mass}
  \pd{J}{\hat{t}} - \frac{1}{\varphi_0^3} \pd{}{\hat{X}} \bracm{\frac{\bracs{J - 1 + \varphi_0}^3}{J^2} \pd{\hat{p}}{\hat{X}}} = 0
\end{equation}
where $\hat{t}$ is the dimensionless time and $\varphi_0$ is the reference porosity. In Eq.~\eqref{mass}, we have considered the dynamic change of permeability through the Kozeny-Carman equation \cite{Choo2018,Zheng2019}. For the initial conditions (ICs) and boundary conditions (BCs), on the top surface ($\hat{X} = 0$), the surface overburden leads to a Dirichlet boundary condition $J = \bar{J}$~\cite{Zheng2019}, while at the bottom ($\hat{X} = 1$), the no-flow condition leads to $\partial J/\partial \hat{X} = 0$. The initial condition is $J = 1$ when $\hat{t} = 0$.

To summarize, we have one governing Eq.~\eqref{mass} and three constitutive laws (with proper ICs \& BCs). Let's index each constitutive law by $i$ ($i=$ 1, 2, or 3). We want to build a model that takes as input $i$ and outputs the solution of Eq.~\eqref{S-V_K}, \eqref{mS-V_K}, or \eqref{N-H} (coupled with Eq.~\eqref{mass} which is common to all). We discuss our approach in the next section.

\section{Multi-Constitutive Neural Network}
\label{PINN}
\begin{figure}[htb]
	\begin{center}
	\includegraphics[width = 0.99\textwidth]{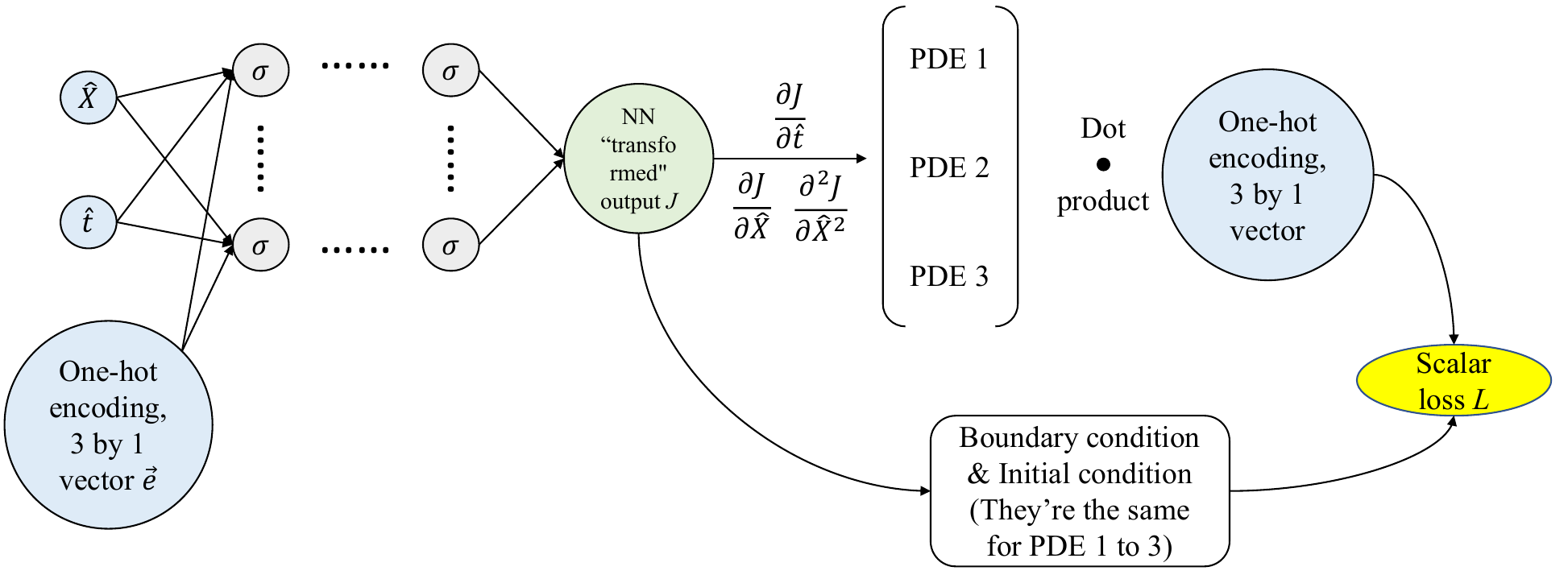}
	\end{center}
	\caption
	{\label{Fig:01}
A schematic diagram of our proposed MCNN. The inputs (left) include $\hat{X}$, $t$ and the law encoding vector $\vec{e}$. The $\sigma$ in the neural network represents the activation function; weights and biases are not shown for simplicity. Time and spatial derivatives of $J$ for the PDE formulations are computed through AD \cite{Baydin2017,Xu2020,Huang2020}.}
\end{figure}

We propose the Multi-Constitutive Neural Network (MCNN) as a novel modeling approach for problems with multiple constitutive laws. As illustrated in Figure~\ref{Fig:01}, we introduce a one-hot constitutive law encoding vector $\vec{e}$ as an extra input to a PINN. Concretely, we use (1, 0, 0), (0, 1, 0), (0, 0, 1) to represent Eqns.~\eqref{S-V_K}, \eqref{mS-V_K}, and \eqref{N-H}, respectively. Therefore, the dimension of each data sample is 5 ($\hat{X}$, $\hat{t}$, and the one-hot encoding vector $\vec{e}$ of length 3). The network output $J$ (after possible final transformation) can be substituted into the three PDEs to calculate the 3 $\times$ 1 residual vector through automatic differentiation (AD) \cite{Baydin2017,Xu2020,Huang2020}. The final scalar PDE loss for one sample is the square of inner product of this residual vector with the same one-hot encoding for this sample. The other loss terms, including boundary and initial conditions closely follow standard PINN (see~\cite{He2020,Berg2018} for details). To be more specific, the scalar loss $\mathcal{L}$ of one training sample could be mathematically expressed as:
\begin{equation}
    \mathcal{L} = \bracl{\sum_{i=1}^3 e_i f_i \bracm{\frac{\partial}{\partial \hat{t} }J\bracs{\hat{X}, \hat{t}, \vec{e}},  \frac{\partial}{\partial \hat{X} }J\bracs{\hat{X}, \hat{t}, \vec{e}}, \frac{\partial^2}{\partial \hat{X}^2 }J\bracs{\hat{X}, \hat{t}, \vec{e}}}}^2 + \mathcal{L}_{BI}
\end{equation}
\begin{equation}
    \mathcal{L}_{BI} = \begin{dcases}
    \bracm{J\bracs{0, \hat{t}, \vec{e}} - \bar{J}}^2 & \qquad \qquad \hat{X} = 0, \quad \hat{t} > 0 \\
    \bracm{\left.\frac{\partial}{\partial \hat{X}} J\bracs{\hat{X}, \hat{t}, \vec{e}}\right\vert_{\hat{X} = 1}}^2 &  \qquad \qquad \hat{X} = 1, \quad  \hat{t} > 0 \\
    \bracm{J\bracs{\hat{X}, 0, \vec{e}} - 1}^2 &  \qquad \qquad \hat{t} = 0 , \quad 0 < \hat{X} < 1 \\
    0 &  \qquad \qquad \hat{t} > 0, \quad 0 < \hat{X} < 1
    \end{dcases}
\end{equation}
where $\mathcal{L}_{BI}$ refers to the loss from the boundary and initial conditions, $e_i$ is the $i^{\rm th}$ component of vector $\vec{e}$, and $f_i\bracm{\cdot, \cdot, \cdot}$ represents the final PDE form for the $i^{\rm th}$ constitutive law, which is obtained by combining Eq.~\eqref{S-V_K}, \eqref{mS-V_K}, or \eqref{N-H} with Eq.~\eqref{mass}.

The material properties and network architectures are presented in Table~\ref{properties} and \ref{parameters}. By leveraging the constitutive law encoding vector, we are able to assimilate multiple constitutive laws and solve all the resulting governing equations through simultaneous training.

\begin{table}[htb]
  \caption{Material properties (same for all three PDEs).}
  \label{properties}
  \centering
  \begin{tabular}{ccccc}
    \toprule
    $\hat{\gamma}$ & $\hat{\mu}$  & $\varphi_0$  & $\bar{J}$ & $\hat{t}$ \\
    \midrule
    1/3 & 1/3  &  0.3 & 0.8 & $0 \leq \hat{t} \leq 1$ \\
    \bottomrule
  \end{tabular}
\end{table}

\begin{table}[htb]
  \caption{Model parameters of the neural networks. The total size of the training set for MCNN is the same as each independent PINN. All architectures use 5 (hidden layers) $\times$ 50 (neurons) with \texttt{tanh} as the activation function. The test sets are always equispaced within the domain. Due to the issue of numerical instability of the Saint-Venant Kirchhoff law \cite{Zheng2019}, we adopt a larger number of epochs for law 1 to make the optimization process more stable. When the number of epochs is fixed, we tune the learning rate and find that the best value is \texttt{5e-4}. These learning rates and numbers of epochs are also typical values used in \cite{He2020,Guo2020,Haghighat2020}.}
  \label{parameters}
  \centering
  \begin{tabular}{llllll}
    \toprule
    DNN    & Training set  & Test set & Optim. & Rate & Epochs \\
    \midrule
    MCNN (ours) & 1000 (per law)  & $10^4$ (per law) & Adam \cite{Kingma2014} & \texttt{5e-4} & $10^5$ \\
    PINN of law 1 & 3000 & $10^4$ & Adam & \texttt{5e-4} & $5\times10^4$ \\
    PINN of law 2 & 3000 & $10^4$ & Adam & \texttt{5e-4} & $2\times10^4$ \\
    PINN of law 3 & 3000 & $10^4$ & Adam & \texttt{5e-4} & $10^4$ \\
    \bottomrule
  \end{tabular}
\end{table}

\section{Results and Discussions}
\label{results}
In this section we evaluate the inference accuracy of MCNN by simultaneously training with three different constitutive laws. To confirm the effectiveness of MCNN, for baseline models, we train an independent PINN respectively for each constitutive law and also calculate the reference solution. To build our DNN structure, we multiply the network output (see Figure~\ref{Fig:01}) with $\hat{X}$ and add $\bar{J}$; this becomes our final output $J$. In this way, our output automatically satisfies the Dirichlet boundary condition at $\hat{X} = 0$. The reference prediction $J$ is computed using a conditionally stable finite difference (FD) algorithm \cite{Zheng2019,Selvadurai2016} with $\Delta \hat{t} = 10^{-5}$ and $\Delta \hat{X} = 0.02$ (second-order accurate in space and first-order accurate in time).

We compute the relative errors to the reference predictions for MCNN and individual PINNs using the formula $\sqrt{{\rm Ave}\bracm{\bracs{J/J_{\rm ref} - 1}^2}}$, as shown in Table~\ref{relative_err}. Surprisingly, MCNN achieves lower relative error than individual PINN on the cases of law 2 and law 3, even with smaller number of training points for each constitutive law. This shows the data efficiency of MCNN from simultaneous training. Since the trend of $J$ with $\hat{X}$ and $\hat{t}$ is fixed (similar to the trend of excess pore pressure in Terzaghi's consolidation equation \cite{Detournay1993}),  the training points for different one-hot encoding vectors can ``communicate'' useful information and benefit from each other. This might explain why we observe better generalization for our method.

\begin{figure}[htb]
  \begin{center}
    \begin{tabular}{ccc}
    \includegraphics[width = 0.3\textwidth]{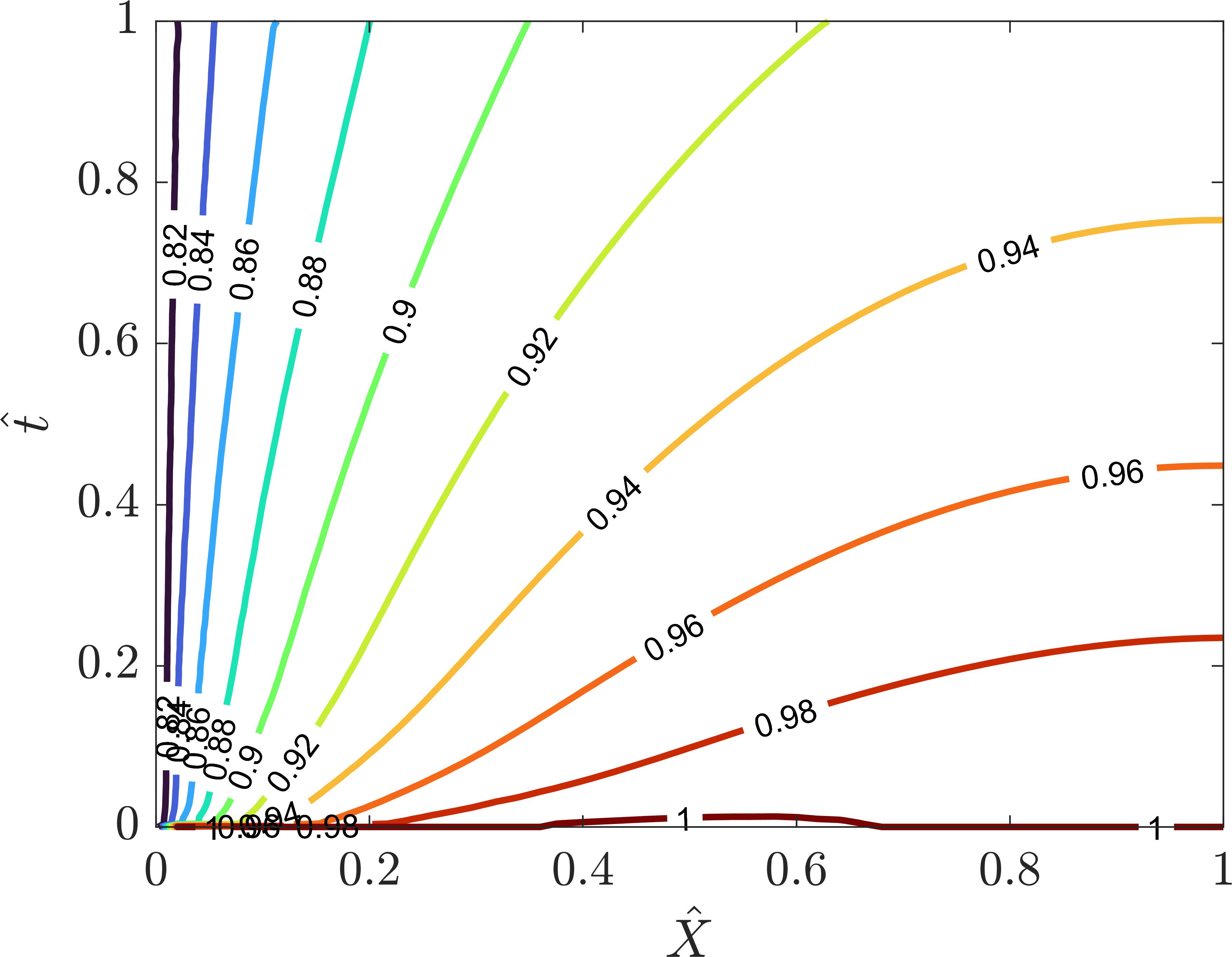} & \includegraphics[width = 0.3\textwidth]{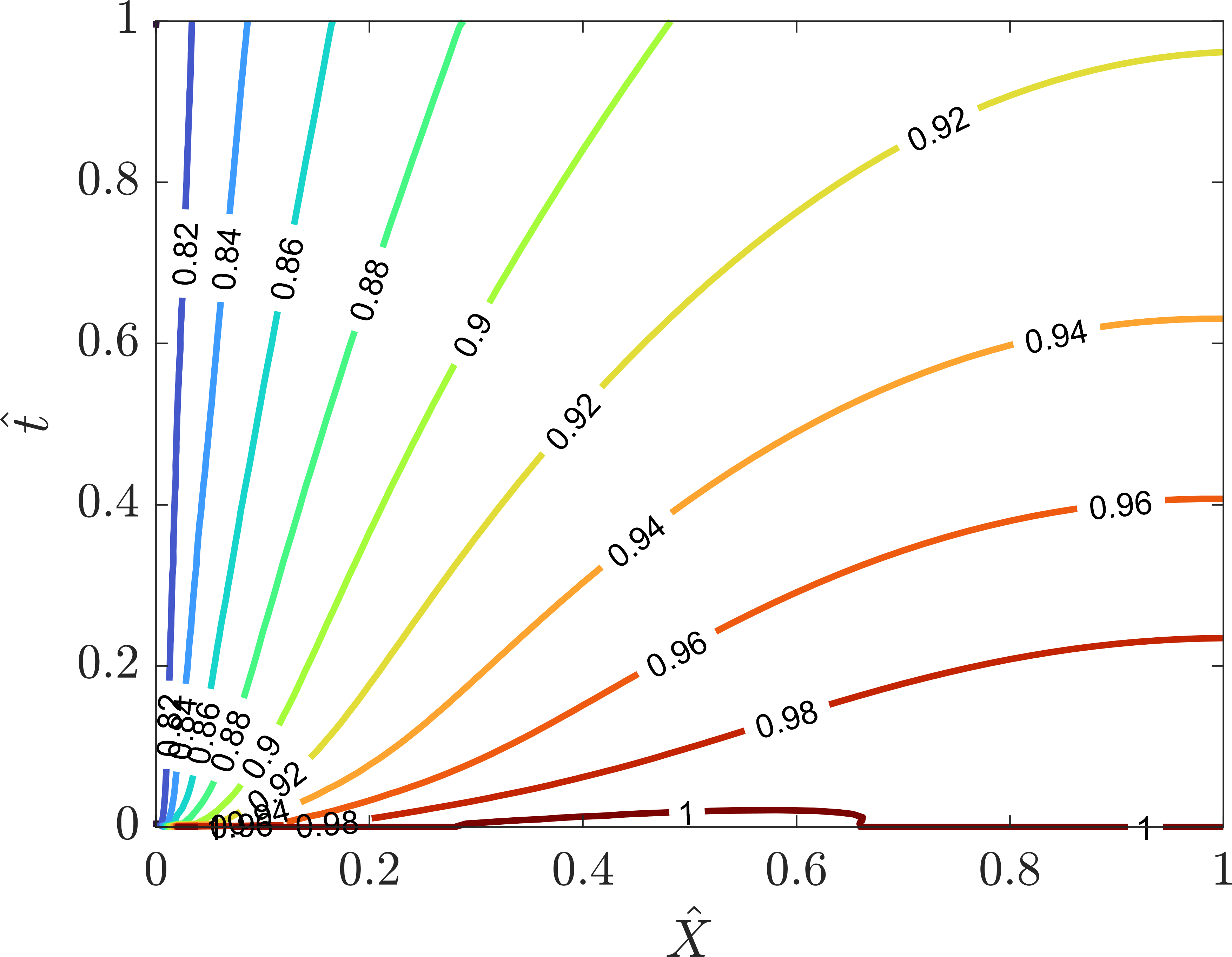} & \includegraphics[width = 0.3\textwidth]{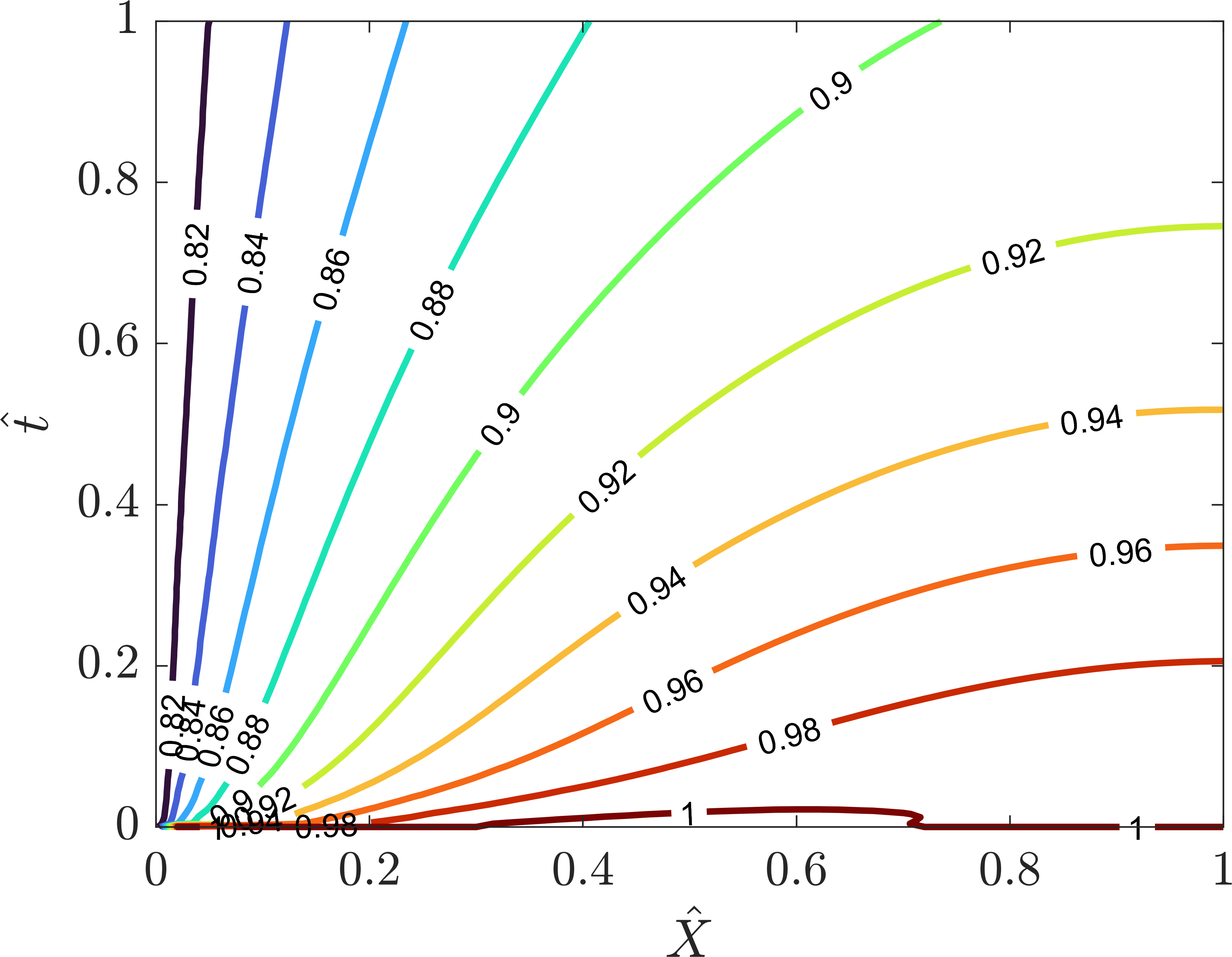} \\
    (a) Law 1, MCNN & (b) Law 2, MCNN & (c) Law 3, MCNN \\
    & & \\
    \includegraphics[width = 0.3\textwidth]{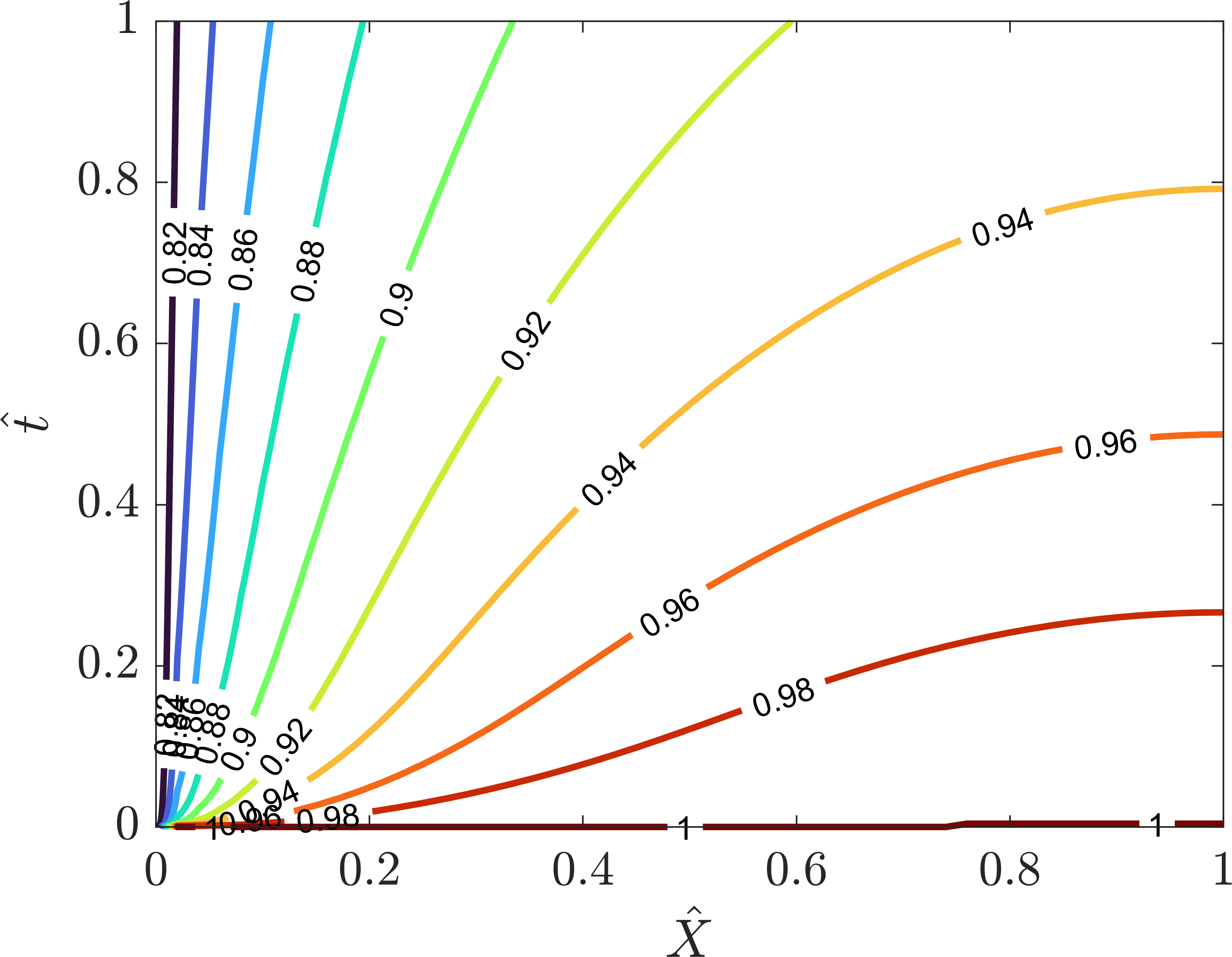} & \includegraphics[width = 0.3\textwidth]{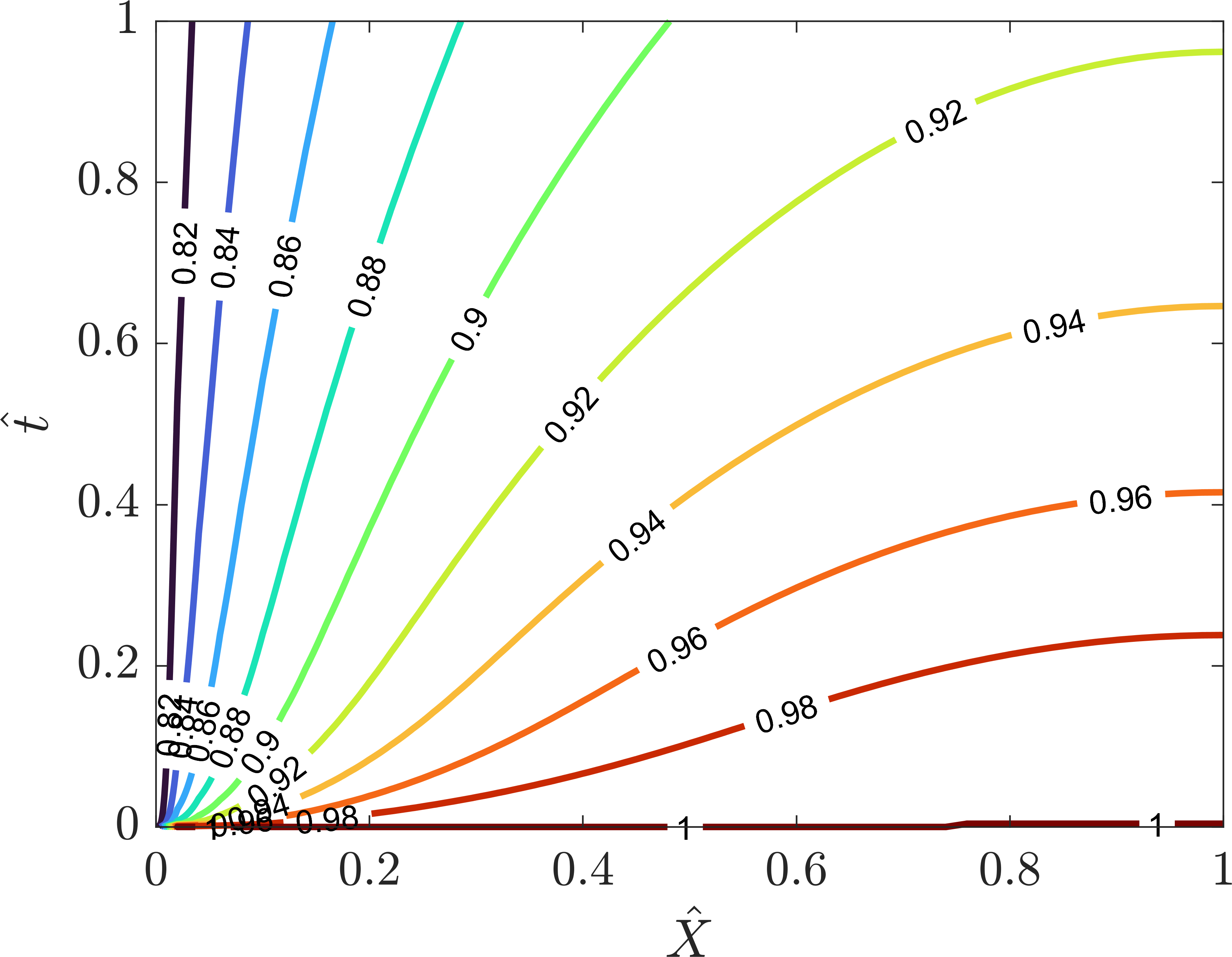} & \includegraphics[width = 0.3\textwidth]{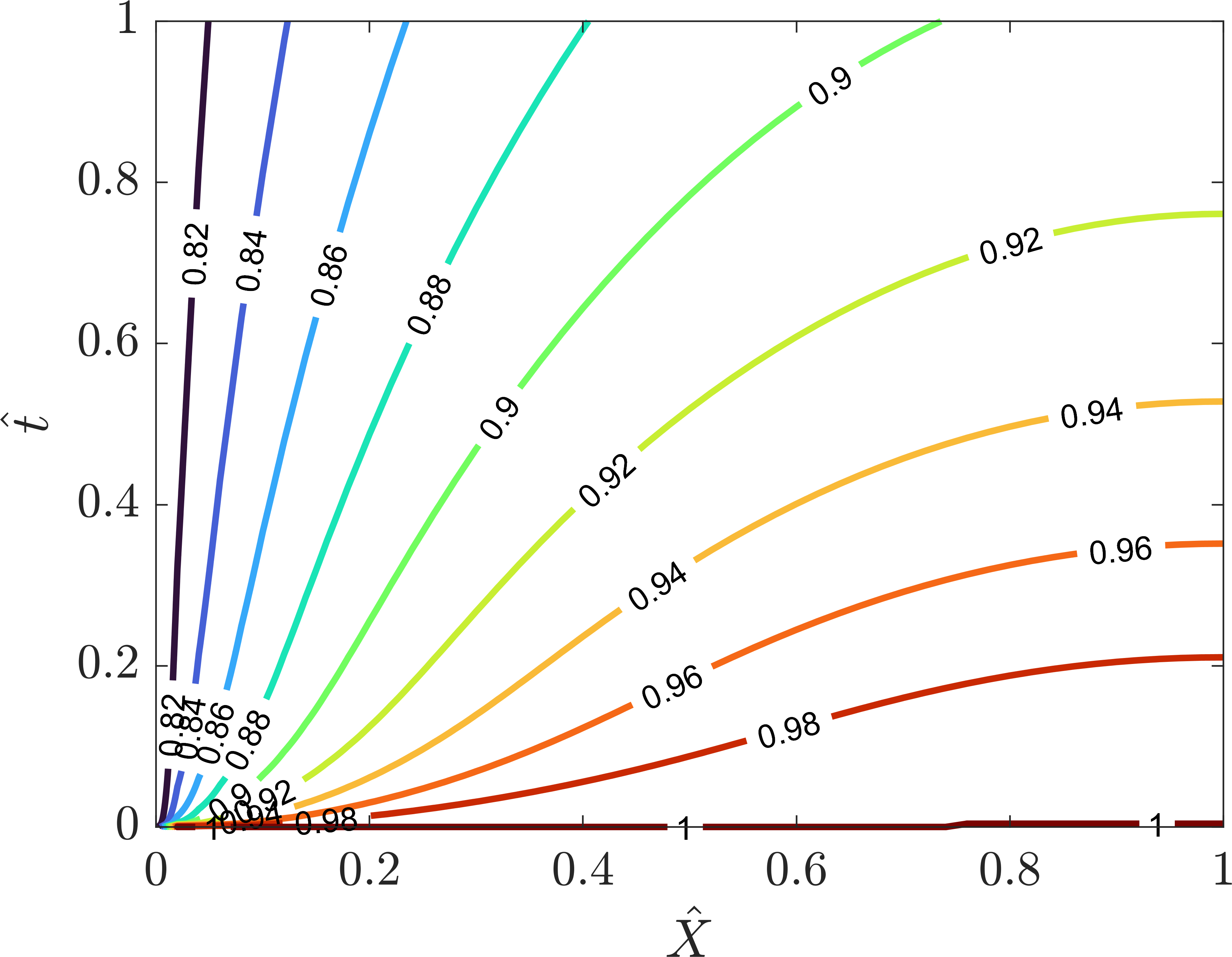} \\
    (d) Law 1, FD & (e) Law 2, FD & (f) Law 3, FD \\
    & & \\
    \includegraphics[width = 0.3\textwidth]{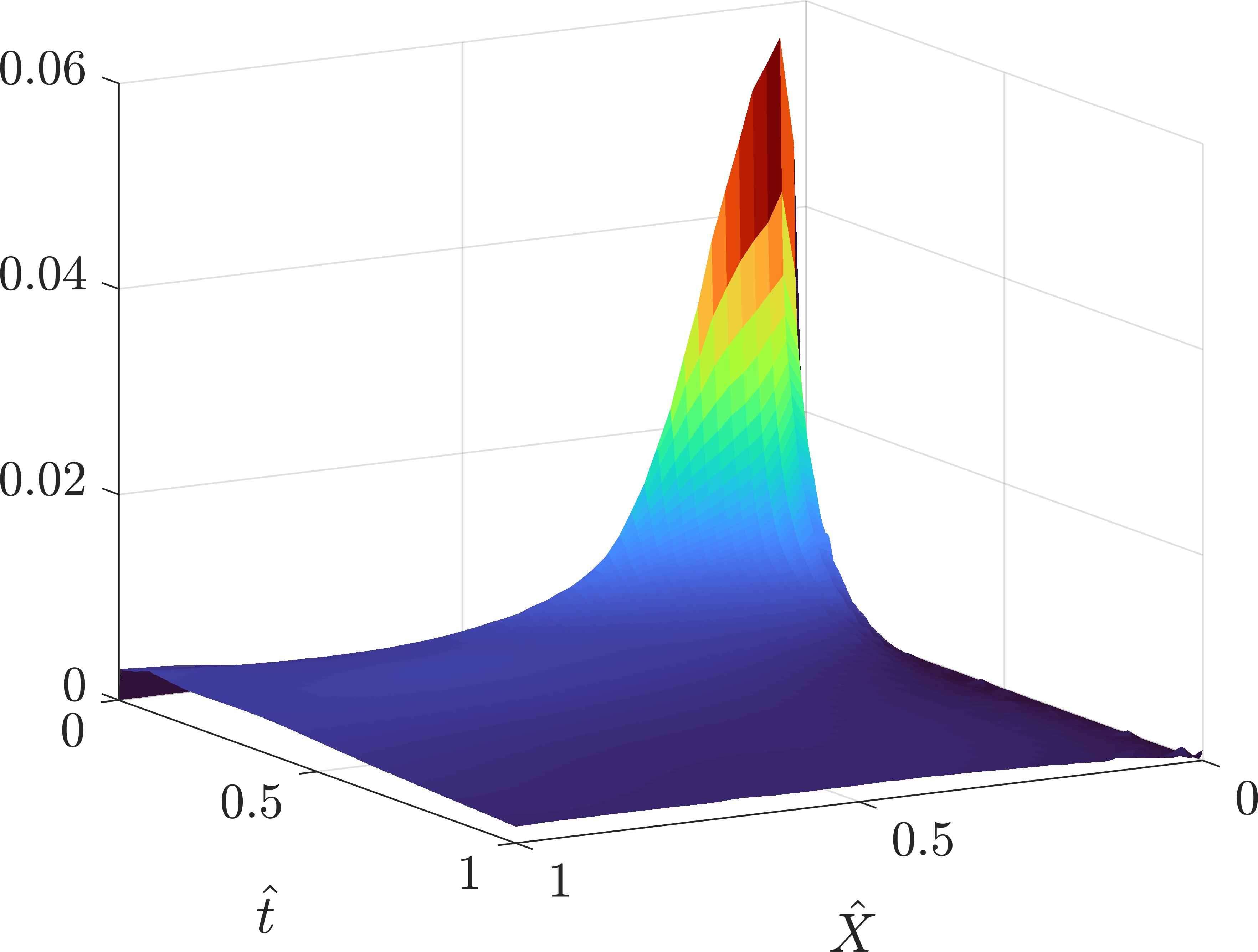} & \includegraphics[width = 0.3\textwidth]{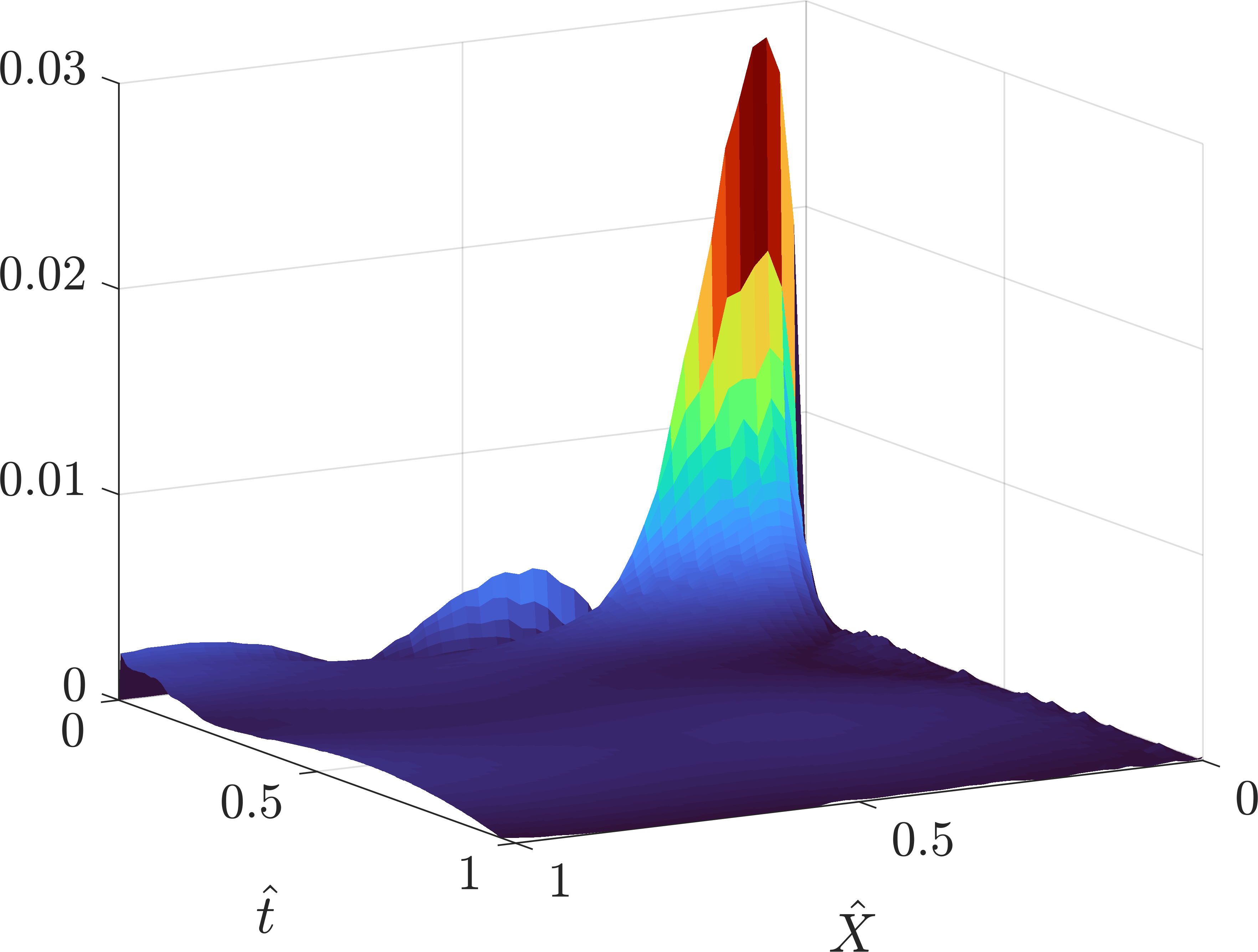} & \includegraphics[width = 0.3\textwidth]{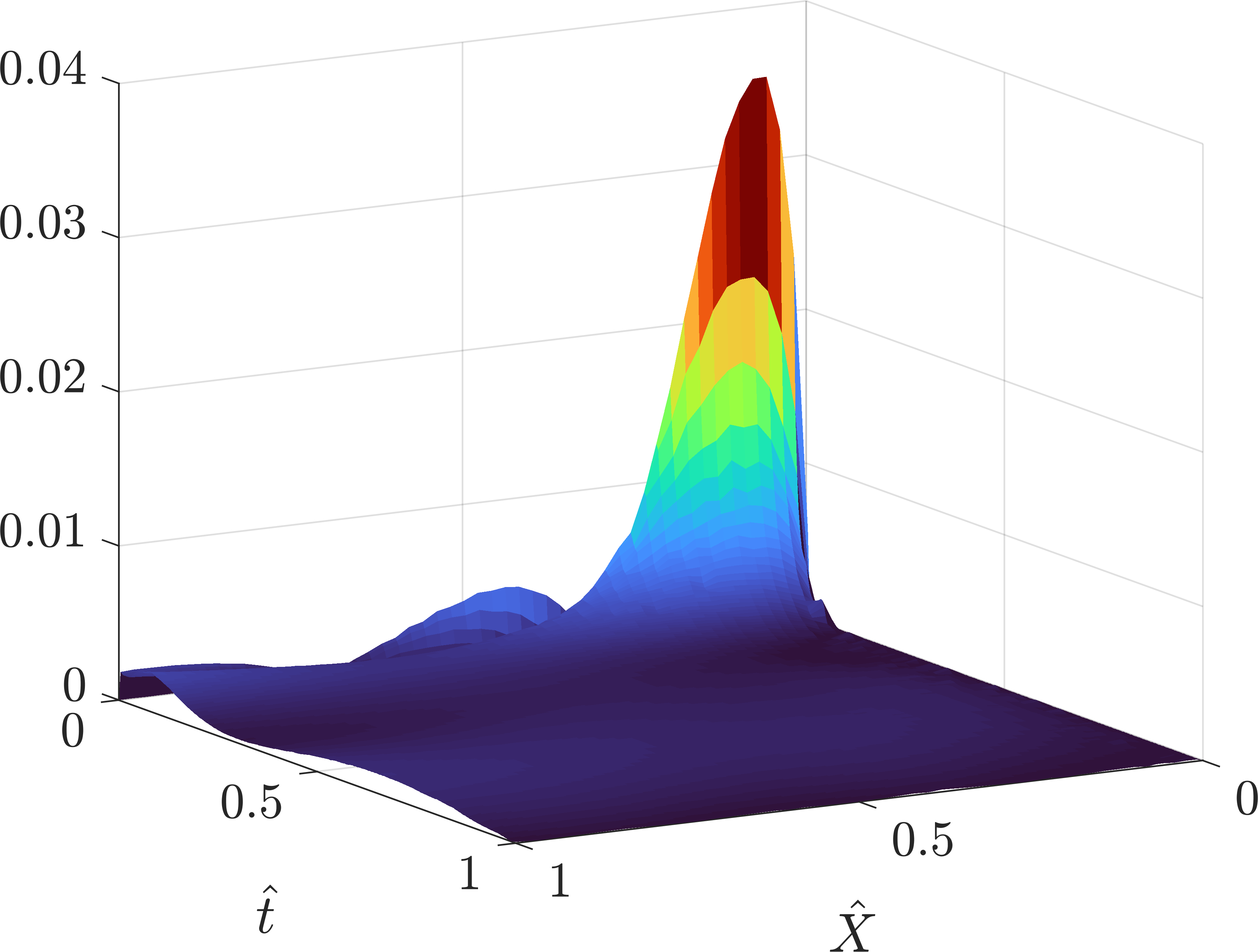} \\
    (g) Law 1, $\abs{\rm MCNN - FD}$ & (h) Law 2, $\abs{\rm MCNN - FD}$ & (i) Law 3, $\abs{\rm MCNN - FD}$
    \end{tabular}
	\end{center}
	\caption
	{\label{Fig:02} Plots of predicted $J$ with MCNN and FD. MCNN is able to make an accurate inference for each constitutive law.}
\end{figure}
\begin{figure}[htb]
  \begin{center}
    \begin{tabular}{c}
\includegraphics[width = 0.65\textwidth]{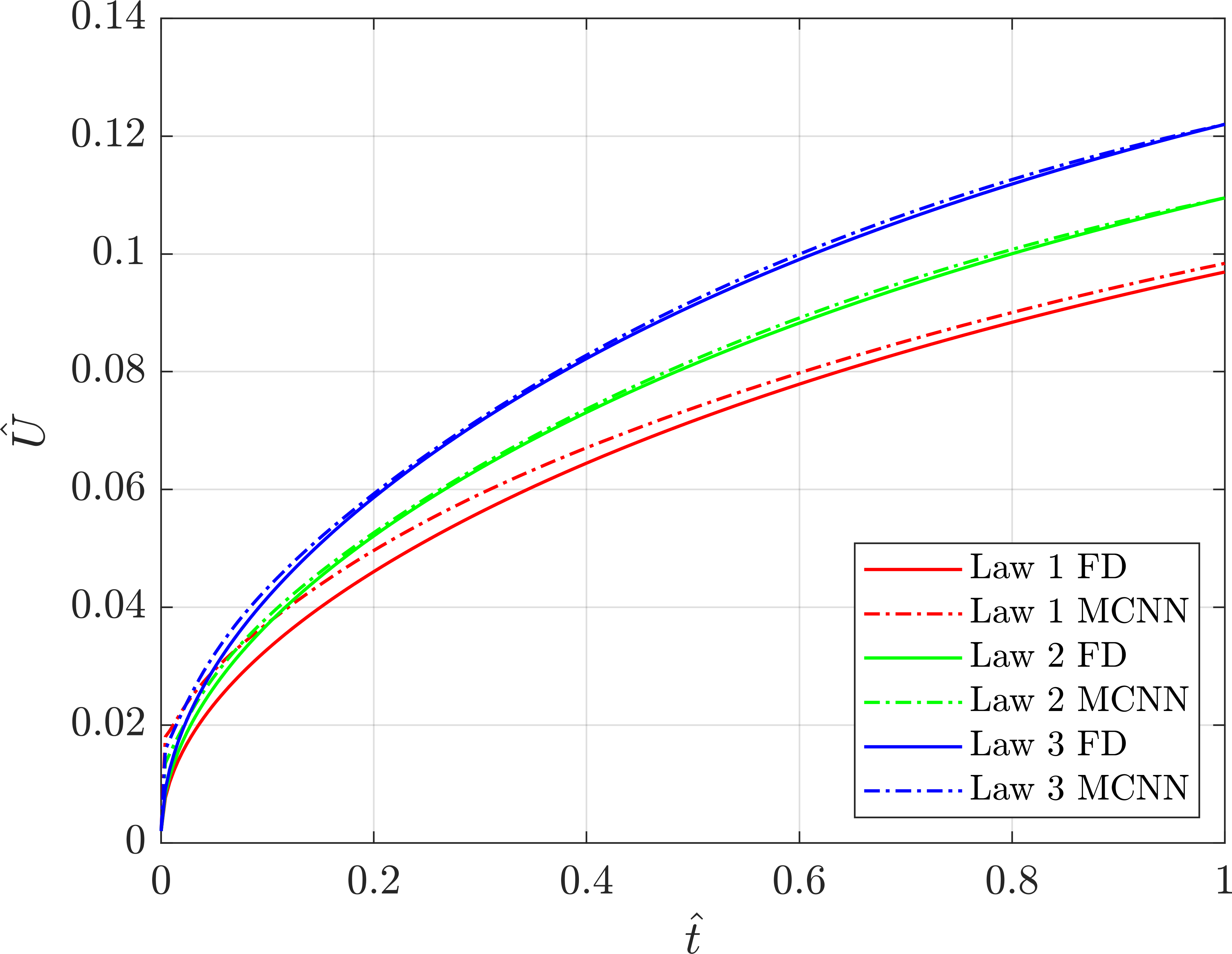}
    \end{tabular}
	\end{center}
	\caption
	{\label{Fig:03}
A visualization of MCNN predictions and reference solutions for the dimensionless settlement $\hat{U}$. Predictions from MCNN, even though they are generated by the same model, are very accurate for all three constitutive laws.}
\end{figure}

In Figure~\ref{Fig:02}, we plot the prediction of $J$ for MCNN and FD versus $\hat{X}$ and $\hat{t}$, from which we can see that MCNN is able to distinguish among three different encoding vectors. In other words, MCNN successfully learns different constitutive laws. In Figure~\ref{Fig:03}, we visualize the dimensionless settlement $\hat{U}$ for each constitutive law. Here $\hat{U}$ is obtained from a trapezoidal numerical integration of $J$ since $J = 1 + \partial \hat{U}/ \partial \hat{X}$ and $\hat{U} \big(\hat{X} = 1\big) = 0$. MCNN achieves accurate predictions for all three constitutive laws, which again demonstrates the powerful capacity of MCNN in modeling the multi-constitutive problem.

\begin{table}[htb]
  \caption{Relative error of the proposed MCNN and PINNs for different constitutive laws. Although MCNN is learning to solve 3 PDEs simultaneously, it outperforms PINNs optimized independently on the cases of law 2 and law 3.}
  \label{relative_err}
  \centering
  \begin{tabular}{lccc}
    \toprule
    Method & Law 1 & Law 2 & Law 3 \\
    \midrule
    MCNN (Ours) & 0.3864\%  & \textbf{0.1322\%}   &  \textbf{0.1665\%}  \\
    Independent PINN & \textbf{0.2936\%}  &  0.2413\%  & 0.4098\% \\
    \bottomrule
  \end{tabular}
\end{table}

\section{Conclusion}
\label{conclude}
In this paper, we propose a novel modeling technique called Multi-Constitutive Neural Network (MCNN) to study multi-constitutive problems. Different from classical PINN, by using a law-encoding input vector, MCNN is able to use a single model to solve for different constitutive laws. Using applications from nonlinear large deformation poromechanics problem, we demonstrated that MCNN could achieve good accuracies. For future work, we plan on studying encoding vectors of the form $(a,b,c)$ with $a+b+c = 1$, $a \ge 0$, $b \ge 0$, $c \ge 0$ (instead of (1,0,0) for example). This is akin to the idea of fractional derivatives~\cite{Li2018}.



\section*{Broader impact}

As described and benchmarked in this paper, our approach opens the path for endowing deep learning with powerful capacity to study poromechanics. Poromechanics deals with the basic interactions between solid and fluid, which is fundamental in reservoir simulations, geothermal engineering, and oil and gas exploration. As a result, we expect MCNN to have an impact in these fields in the future. For example, in reservoir simulation, by using MCNN, we could compare the differences between Darcy and non-Darcy flows using a single training process. This research may have some negative impacts on society because it may encourage over-exploitation of oil and gas sites, pollution of sites (e.g., fracking), and may lead to an overall increase in production and consumption of non-renewable energy sources, contributing for example to global warming.

\end{document}